\documentclass[letterpaper]{article}
\usepackage{aaai2027}
\usepackage[hyphens]{url}
\usepackage{graphicx}
\usepackage{natbib}
\usepackage{caption}
\usepackage{amsmath}
\usepackage{amssymb}
\usepackage{array}
\usepackage{multirow}
\usepackage{algorithm}
\usepackage{algorithmic}
\usepackage{booktabs}
\usepackage[table]{xcolor}
\usepackage{xspace}

\definecolor{CaptionBlue}{RGB}{226,236,250}
\definecolor{VQAPink}{RGB}{246,218,240}
\definecolor{LastColBlue}{RGB}{226,236,250}
\definecolor{DropTeal}{RGB}{0,120,115}

\newcommand{\best}[1]{\textbf{#1}}

\newcommand{\method}{QR-STT\xspace}
\newcommand{\methodfull}{QR-Structured Thermal Triggers\xspace}

\title{QR-Structured Thermal Triggers for Targeted Semantic Attacks on Infrared Vision--Language Models}
\author{Xiang Chen, Yingying Zhao, Chao Li, Jiaju Han, Ben Zhang, Ang Li, Jiahuan Long, Yiwei Wei, Jiujiang Guo, Chengyin Hu}
\affiliations{}
\nocopyright

\begin{document}
\maketitle

\begin{abstract}
Infrared vision--language models (IR-VLMs) extend thermal perception from closed-set recognition to open-vocabulary classification, image captioning, and visual question answering (VQA), enabling infrared inputs to be interpreted through language-aligned representations. However, their robustness to structured thermal perturbations and the stability of cross-modal semantic alignment remain insufficiently studied. In this paper, we introduce QR-Structured Thermal Triggers (\method), a stealthy, training-free black-box framework for targeted semantic steering against IR-VLMs. \method constructs a low-contrast thermal trigger with preserved QR functional regions and optimizable internal modules, where each module is assigned a cold, neutral, or hot thermal state. Within this interpretable module space, \method jointly optimizes module topology and rendering parameters, including placement, scale, rotation, intensity, blur, and roundness, to adapt the trigger to infrared imagery while maintaining visual stealth. A three-stage gradient-free search with greedy module-flip refinement handles the mixed discrete--continuous attack space. The objective promotes target alignment, suppresses source evidence, and regularizes QR structure and visual stealth, enabling source-to-target steering without model training, data poisoning, or gradient access. Experiments across multiple CLIP-style encoders show that \method reliably redirects image--text alignment toward attacker-chosen concepts while maintaining visual stealth. Moreover, adversarial images optimized for classification transfer to captioning and VQA tasks, showing that CLIP-side steering can propagate to generation-level behavior and induce target-consistent semantic drift. These findings reveal QR-structured thermal triggers as an interpretable attack surface for language-driven infrared perception and motivate robustness evaluation against structured cross-task semantic risks.
\end{abstract}

\section{Introduction}

\begin{figure}[t]
\centering
\includegraphics[width=0.48\textwidth]{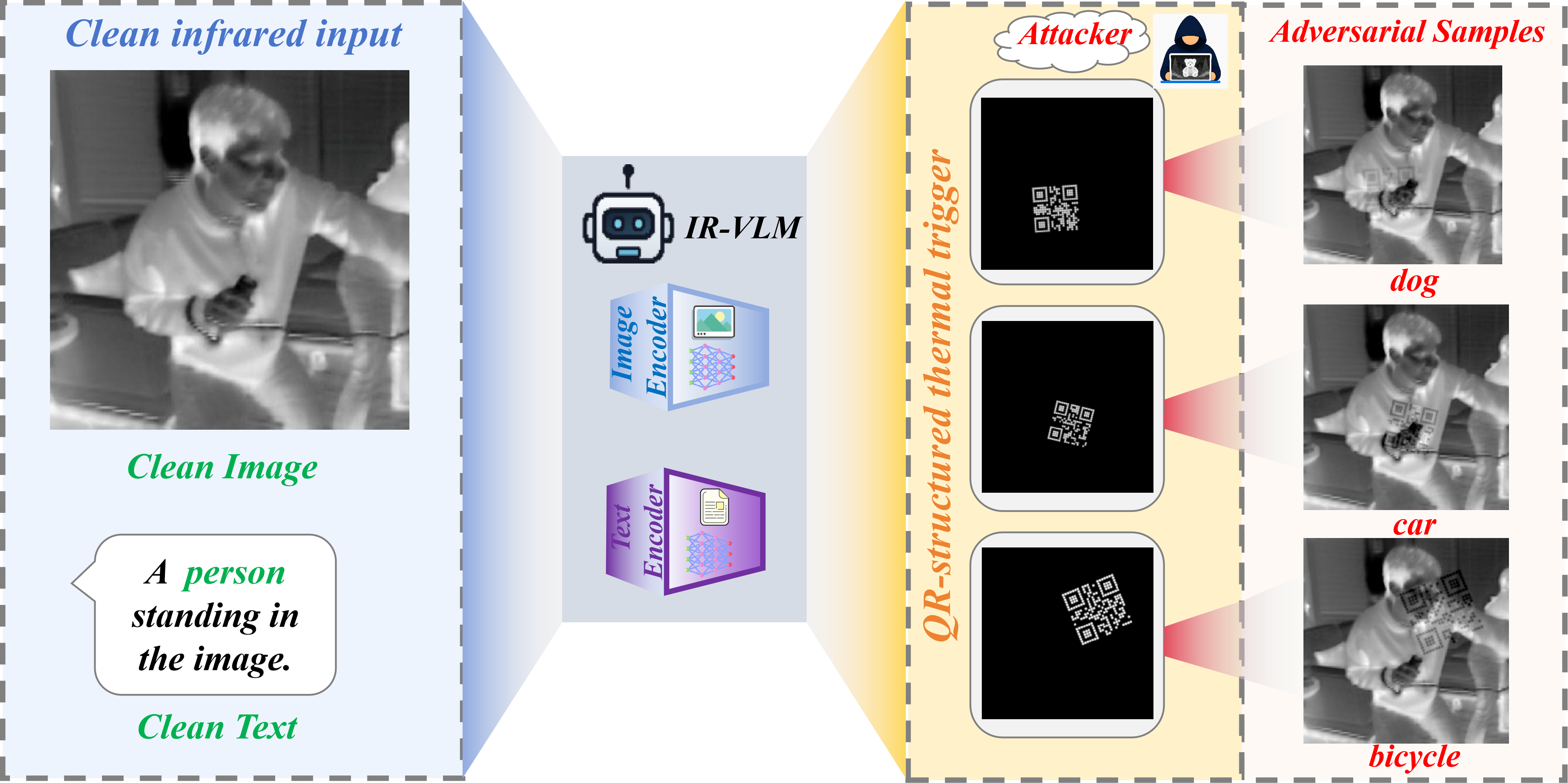}
\caption{A QR-structured trigger redirects the image--text alignment of a clean infrared input toward attacker-specified concepts such as dog, car, and bicycle.}
\label{fig:intro}
\end{figure}

Vision--language models (VLMs) align visual and linguistic representations through large-scale cross-modal pre-training, supporting open-vocabulary classification, image captioning, and visual question answering (VQA)~\citep{radford2021clip,cherti2023openclip,xu2024metaclip,sun2023evaclip,liu2023llava,liu2024llavanext,li2023blip2,dai2023instructblip}. Recent infrared vision--language models (IR-VLMs) extend this capability to thermal imagery, enabling infrared scenes to be interpreted through language queries rather than fixed detector categories~\citep{jiang2024infraredllava,cao2025irgpt,zhang2025ifbench}. Such models are promising for nighttime driving, surveillance, search-and-rescue, and adverse-weather perception, where thermal sensing remains informative under poor illumination.

This transition from closed-set recognition to language-aligned perception introduces a distinct security risk. Conventional infrared detectors operate over predefined labels, and existing attacks typically aim to suppress detections or induce generic errors. IR-VLMs instead compare visual inputs with a broad set of textual concepts in a shared embedding space. A compact adversarial cue may therefore increase alignment with an attacker-chosen concept without fully removing the original visual evidence. This targeted semantic steering is more consequential than untargeted degradation because the attacker controls the direction of the error. As shown in Figure~\ref{fig:intro}, the same infrared input can be redirected toward concepts such as dog, car, or bicycle.

Existing robustness studies do not fully characterize this threat. Most VLM attacks are developed in the visible spectrum and rely on pixel perturbations, transferable adversarial examples, patches, prompt-related vulnerabilities, illumination transformations, or general multimodal attacks~\citep{lu2023setlevel,yin2023vlattack,zhao2023evaluating,zhang2025anyattack,xie2025chain,liu2025lighting,nie2026vattack,hu2026omniattack,guo2026physpatch}. Infrared adversarial research has primarily targeted detectors and trackers using physical patches, stickers, grids, hot/cold blocks, or curve-based thermal patterns~\citep{wei2023physically,zhu2024infrared,tiliwalidi2025advgrid,hu2023hotcold,hu2024advib,hu2024infraredcurves,jia2025advicrs}. These methods mainly pursue missed detections, false positives, or general performance degradation. Whether a compact structured carrier can deliberately manipulate the open-vocabulary semantics of IR-VLMs remains underexplored.

We investigate QR codes as structured attack carriers in the digital infrared image domain. A QR code naturally separates fixed functional regions from editable internal modules. Finder patterns, timing patterns, format-related regions, and the quiet zone preserve its global organization, while the remaining modules define a discrete topology that can be optimized for target-specific steering. Unlike unconstrained patches, this representation explicitly separates the preserved scaffold from the attack variables, providing a structured and interpretable space for encoding adversarial cues.

A direct high-contrast QR overlay would be visually conspicuous and methodologically similar to ordinary patch insertion. We therefore assign each editable module one of three relative infrared intensity states: cold, neutral, or hot, corresponding to local intensity decreases, unchanged responses, and local intensity increases. The trigger is further controlled by trigger scale, perturbation intensity, active-module ratio, and structural similarity. This yields a low-contrast QR-structured perturbation while retaining the native QR organization. Our central question is whether the functional scaffold and editable module topology of a QR code can jointly support targeted semantic steering under explicit visual-distortion constraints.

Based on this formulation, we propose \methodfull (\method), a training-free black-box targeted attack for IR-VLMs. Given a clean infrared image and an attacker-specified target label, \method jointly optimizes trigger placement, scale, rotation, rendering parameters, and internal module states. Its objective promotes target alignment, suppresses source evidence, regularizes QR topology, and limits visual distortion. To address the resulting mixed discrete--continuous search space, we develop a progressive gradient-free strategy comprising coarse zone search, module-topology search, infrared rendering refinement, and greedy module-flip refinement.

We evaluate \method on targeted zero-shot classification across multiple CLIP-style encoders and further assess transfer to image captioning and VQA. The downstream generative models are excluded from attack optimization. Consequently, target-consistent changes in their outputs indicate that the induced alignment shift extends beyond a task-specific classifier and propagates to generation-level behavior.

Our contributions are summarized as follows:
\begin{itemize}
    \item We identify attacker-directed semantic steering as a distinct robustness risk for IR-VLMs and introduce a QR-structured attack framework that targets open-vocabulary image--text alignment rather than generic recognition failure.

    \item We formulate QR trigger generation as a constrained mixed discrete--continuous optimization problem that separates fixed functional regions from editable cold, neutral, and hot modules, and develop a progressive gradient-free solver for joint topology and rendering optimization.

    \item We conduct comprehensive evaluations across multiple CLIP-style encoders, image-captioning models, and VQA models. Comparisons with structured infrared baselines and detailed ablations demonstrate the effectiveness, cross-task transferability, and interpretability of \method.
\end{itemize}

\section{Related Work}

\textbf{Infrared vision--language models.}
Vision--language models align images and text through large-scale contrastive or generative pretraining \citep{radford2021clip,li2023blip2,dai2023instructblip,zhai2023siglip}. Recent infrared and thermal VLM studies adapt these ideas to low-texture thermal imagery, where discriminative cues depend more on heat radiation, object silhouette, and sensor statistics than RGB color or fine-grained texture \citep{jiang2024infraredllava,cao2025irgpt,moshtaghi2025rgbth,zhang2025ifbench}. These models make infrared perception more language-accessible, but also introduce open-vocabulary semantic attack surfaces.

\begin{figure*}[t]
\centering
\includegraphics[width=0.98\textwidth]{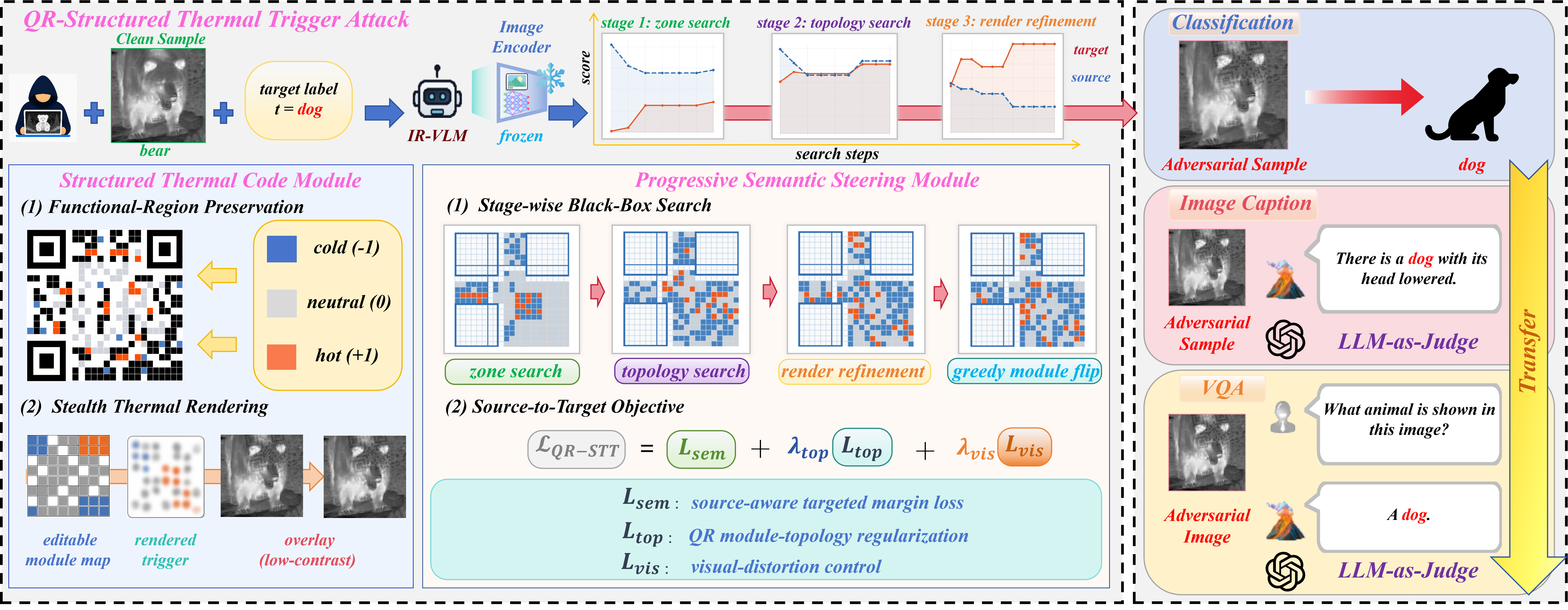}
\caption{Overview of \method. Given a clean infrared image and an attacker-specified target label, \method constructs a QR-structured thermal trigger with fixed functional regions and editable cold/neutral/hot modules. The attack progressively optimizes the trigger through zone search, topology search, thermal rendering refinement, and greedy module flipping, steering a frozen IR-VLM toward the target concept and transferring the induced semantic drift to classification, captioning, and VQA.}
\label{fig:method}
\end{figure*}

\textbf{Robustness of Vision--Language Alignment under Adversarial Attacks.}
Multimodal adversarial attacks often exploit the shared image--text representation of CLIP-style models, either by perturbing the image, guiding transfer across models, or inducing downstream caption/VQA errors \citep{lu2023setlevel,yin2023vlattack,zhao2023evaluating,zhang2025anyattack,xie2025chain,liu2025lighting,nie2026vattack,hu2026omniattack,guo2026physpatch}. Because instruction-tuned VLMs frequently inherit visual features from CLIP-like encoders \citep{liu2023llava,liu2024llavanext,awadalla2023openflamingo}, a shift in image--text alignment can propagate to language outputs. Our work focuses on this transfer in infrared scenarios and asks whether a stealthy symbolic thermal carrier can steer the shared representation toward a specified target.

\textbf{Structured infrared perturbations.}
Infrared attacks often constrain perturbations to physically meaningful or low-frequency patterns, including adversarial infrared patches, car stickers, wearable hot/cold blocks, grid patterns, and spline or curve carriers \citep{brown2017adversarial,wei2023physically,zhu2024infrared,hu2023hotcold,hu2024advib,tiliwalidi2025advgrid,hu2024infraredcurves,jia2025advicrs}. These studies mainly target detectors, whereas \method targets open-vocabulary IR-VLM semantics. We adapt representative structured infrared carriers as baselines under the same source-to-target protocol.

\textbf{Symbolic visual carriers and QR structure.}
Standard QR codes have inherent fixed functional regions and editable data modules, making them a highly useful symbolic layout for studying structured visual perturbations \citep{iso18004qrcode}. However, directly pasting a plain QR-like patch is visually obvious and methodologically close to a generic patch attack \citep{brown2017adversarial}. \method instead fully preserves QR functional regions and optimizes only the internal thermal modules under strict stealth constraints, testing whether symbolic module topology itself can serve as an interpretable attack space.

\section{Method}

Figure~\ref{fig:method} illustrates the overall pipeline of \method.
Given an infrared image and an attacker-specified target label,
\method constructs a QR-structured trigger and jointly optimizes its
geometry, editable module topology, and rendering parameters to steer
a frozen CLIP-style IR-VLM toward the target concept. We first
formulate the targeted attack, then introduce the constrained QR
representation, and finally describe the progressive black-box
optimization strategy.

\subsection{Problem Setup}

Let $x_i\in[0,1]^{H\times W}$ denote an infrared image with source
label $s_i$, and let $t\in\mathcal{Y}$ be an attacker-specified target
label, where $t\neq s_i$. We consider a frozen CLIP-style IR-VLM with
an image encoder $f_I(\cdot)$ and a text encoder
$f_T(\cdot)$~\citep{radford2021clip,cherti2023openclip,
xu2024metaclip,sun2023evaclip}. The attacker can query model outputs
but has no access to model parameters or gradients.

For each class $c\in\mathcal{Y}$, we construct a normalized text
prototype from a prompt set $\mathcal{T}_c=\{T_c^k\}_{k=1}^{K}$:
\begin{equation}
z_c=
\operatorname{norm}
\left(
\frac{1}{K}
\sum_{k=1}^{K}f_T(T_c^k)
\right).
\label{eq:text_proto}
\end{equation}
The image--text similarity score is:
\begin{equation}
S_c(x)=
\left\langle
\operatorname{norm}(f_I(x)),z_c
\right\rangle .
\label{eq:score}
\end{equation}
Here $\operatorname{norm}(\cdot)$ denotes $\ell_2$ normalization and
$\langle\cdot,\cdot\rangle$ denotes cosine similarity. The attack
succeeds when the target becomes the top-ranked class:
\begin{equation}
\arg\max_{c\in\mathcal{Y}}
S_c(x_i^{\mathrm{adv}})=t .
\label{eq:success}
\end{equation}

The adversarial image is generated by a QR-structured rendering
operator:
\begin{equation}
x_i^{\mathrm{adv}}
=
A_{\mathrm{QR}}(x_i;\theta),
\label{eq:attack_map}
\end{equation}
where
$\theta=\{g,h,M_{\mathrm{edit}}\}$.
The geometry
$g=(c_x,c_y,s,r)$ contains the trigger center, scale, and rotation,
while the rendering parameters
$h=(\alpha,\sigma,\rho)$ control intensity, blur radius, and module
roundness. $M_{\mathrm{edit}}$ denotes the states of the editable QR
modules.

\subsection{QR-Structured Thermal Trigger}

A direct high-contrast QR overlay is visually conspicuous and
methodologically similar to ordinary patch insertion. Instead,
\method exploits the native organization of a QR code to define a
constrained module space. We partition the QR grid into fixed
functional regions $\mathcal{F}$ and editable internal regions
$\mathcal{E}$. The fixed regions contain finder patterns, timing-like
modules, format-like regions, and a quiet-zone-like boundary,
preserving the global QR scaffold. The editable regions provide the
degrees of freedom for target-specific optimization.

Let $M^0$ denote the original QR template and $M$ the optimized
module map. We impose the hard constraint:
\begin{equation}
\begin{cases}
M_{uv}=M^0_{uv}, & (u,v)\in\mathcal{F},\\
M_{uv}\in\{-1,0,+1\}, & (u,v)\in\mathcal{E}.
\end{cases}
\label{eq:module_state}
\end{equation}
The values $-1$, $0$, and $+1$ represent cold, neutral, and hot
states, corresponding to local intensity decreases, unchanged
responses, and local intensity increases in the normalized infrared
image domain.

Directly searching over ternary states leads to a high-dimensional
discrete problem. We therefore associate each editable module with a
relaxed variable $q_{uv}$ and decode it using two thresholds:
\begin{equation}
M_{uv}=
\begin{cases}
-1, & q_{uv}<\tau_{-},\\
0,  & \tau_{-}\le q_{uv}\le\tau_{+},\\
+1, & q_{uv}>\tau_{+}.
\end{cases}
\label{eq:threshold}
\end{equation}
Here $\tau_{-}$ and $\tau_{+}$ denote the lower and upper decoding
thresholds. This relaxation enables continuous gradient-free search
while preserving a discrete final topology.

Given the decoded module map, \method constructs a signed intensity
response and embeds it into the infrared image:
\begin{equation}
x_i^{\mathrm{adv}}
=
\operatorname{clip}
\left(
x_i+
\mathcal{W}
\big(
R(M_{\mathrm{edit}};h);g
\big),
0,1
\right).
\label{eq:render}
\end{equation}
Here $R(\cdot)$ converts the module states into a signed infrared
intensity map, and $\mathcal{W}(\cdot)$ applies translation, scaling,
and rotation. Hot modules increase local intensity, cold modules
decrease it, and neutral modules remain inactive. Blur and rounded
module boundaries suppress sharp digital artifacts and improve
consistency with infrared image characteristics.

\subsection{Source-to-Target Objective}

We formulate trigger generation using three objectives with distinct
roles:
\begin{equation}
\mathcal{L}_{\method}
=
L_{\mathrm{sem}}
+
\lambda_{\mathrm{top}}L_{\mathrm{top}}
+
\lambda_{\mathrm{vis}}L_{\mathrm{vis}} .
\label{eq:total_loss}
\end{equation}
Here $L_{\mathrm{sem}}$ drives targeted semantic steering,
$L_{\mathrm{top}}$ regularizes the editable module topology, and
$L_{\mathrm{vis}}$ constrains image-level distortion. The three terms
address complementary requirements and avoid redundant optimization
objectives.

\paragraph{Targeted semantic steering.}
A targeted attack requires the target score to exceed both the source
score and all remaining competing classes. We therefore define:
\begin{equation}
\begin{aligned}
L_{\mathrm{sem}}
&=
\left[
\max_{c\in\mathcal{Y}\setminus\{t,s_i\}}
S_c(x_i^{\mathrm{adv}})
-
S_t(x_i^{\mathrm{adv}})
+
m
\right]_+
\\
&\quad+
\beta
\left[
S_{s_i}(x_i^{\mathrm{adv}})
-
S_t(x_i^{\mathrm{adv}})
+
m_s
\right]_+ ,
\end{aligned}
\label{eq:semantic_loss}
\end{equation}
where $[u]_+=\max(u,0)$. The first term separates the target from
the strongest non-source competitor with margin $m$, while the second
explicitly separates the target from the source concept with margin
$m_s$. The coefficient $\beta$ controls the strength of source
suppression. This unified objective directly encodes the conditions
required for source-to-target steering without introducing
overlapping classification losses.

\paragraph{Module-topology regularization.}
The QR functional regions are already preserved by the hard constraint
in Eq.~\eqref{eq:module_state}. We therefore regularize only the
editable topology:
\begin{equation}
L_{\mathrm{top}}
=
L_{\mathrm{act}}
+
\eta_{\mathrm{tv}}L_{\mathrm{tv}} .
\label{eq:topology_loss}
\end{equation}

The active-module term controls the proportion of cold and hot
modules:
\begin{equation}
L_{\mathrm{act}}
=
\left|
\frac{1}{|\mathcal{E}|}
\sum_{(u,v)\in\mathcal{E}}
|M_{uv}|
-
\rho_0
\right|,
\label{eq:active_loss}
\end{equation}
where $\rho_0$ denotes the desired active-module ratio. Since
$|M_{uv}|=1$ for cold or hot modules and $|M_{uv}|=0$ for neutral
modules, this term prevents both insufficiently expressive sparse
patterns and visually dominant dense patterns.

The spatial regularizer is:
\begin{equation}
L_{\mathrm{tv}}
=
\frac{1}{|\mathcal{E}|}
\sum_{(u,v)\in\mathcal{E}}
\left(
|M_{u+1,v}-M_{uv}|
+
|M_{u,v+1}-M_{uv}|
\right),
\label{eq:tv_loss}
\end{equation}
where invalid boundary terms are omitted. This term discourages
isolated state changes and fragmented checkerboard patterns while
retaining sufficient flexibility for target-specific topology
optimization.

\paragraph{Visual-distortion control.}
We constrain image-level distortion using the mean absolute
perturbation and structural similarity:
\begin{equation}
\begin{aligned}
L_{\mathrm{vis}}
&=
\left[
D(x_i,x_i^{\mathrm{adv}})
-
\delta_{\max}
\right]_+
\\
&\quad+
\mu_s
\left[
\gamma_{\min}
-
\mathrm{SSIM}(x_i,x_i^{\mathrm{adv}})
\right]_+ ,
\end{aligned}
\label{eq:vis_loss}
\end{equation}
where
\begin{equation}
D(x_i,x_i^{\mathrm{adv}})
=
\frac{1}{HW}
\left\|
x_i^{\mathrm{adv}}-x_i
\right\|_1
\label{eq:distortion}
\end{equation}
denotes the mean absolute perturbation.
$\delta_{\max}$ specifies the maximum allowed perturbation and
$\gamma_{\min}$ the minimum structural similarity (SSIM) ~\citep{wang2004ssim}. The hinge form penalizes only violations of
the corresponding visual constraints. Trigger coverage is not
penalized separately because it is already controlled by the trigger
scale and the active-module ratio.

\begin{table*}[t]
\centering
\small
\caption{Zero-shot classification results. We report attack success rate (\%) for each target category and the macro-average over all model--target pairs.}
\label{tab:main}
\resizebox{0.98\textwidth}{!}{
\begin{tabular}{lccccccccccccc}
\toprule
\multirow{2}{*}{Method}
& \multicolumn{3}{c}{OpenCLIP ViT-B/16}
& \multicolumn{3}{c}{Meta-CLIP ViT-L/14}
& \multicolumn{3}{c}{EVA-CLIP ViT-G/14}
& \multicolumn{3}{c}{OpenAI CLIP ViT-L/14}
& \multirow{2}{*}{Avg.} \\
\cmidrule(lr){2-4}
\cmidrule(lr){5-7}
\cmidrule(lr){8-10}
\cmidrule(lr){11-13}
& Bicycle & Car & Dog
& Bicycle & Car & Dog
& Bicycle & Car & Dog
& Bicycle & Car & Dog
& \\
\midrule

AdvICRS
& 4.97 & 9.05 & 4.13
& 3.60 & 8.30 & 6.40
& 4.52 & 3.15 & 2.31
& 2.80 & 5.60 & 4.40
& 4.94 \\

HCB
& 3.30 & 10.67 & 11.50
& 5.65 & 5.20 & 13.60
& 7.52 & 6.80 & 7.40
& 9.35 & 4.40 & 17.60
& 8.58 \\

AdvGrid
& 8.13 & 14.20 & 11.16
& 15.30 & 11.52 & 19.20
& 10.40 & 17.10 & 15.80
& 6.80 & 11.20 & 25.60
& 13.87 \\

\method
& \best{42.85} & \best{24.79} & \best{38.50}
& \best{35.91} & \best{31.40} & \best{30.27}
& \best{28.40} & \best{35.20} & \best{34.80}
& \best{33.94} & \best{42.95} & \best{32.34}
& \best{34.28} \\

\bottomrule
\end{tabular}}
\end{table*}

\begin{figure*}[t]
\centering
\includegraphics[width=0.98\textwidth]{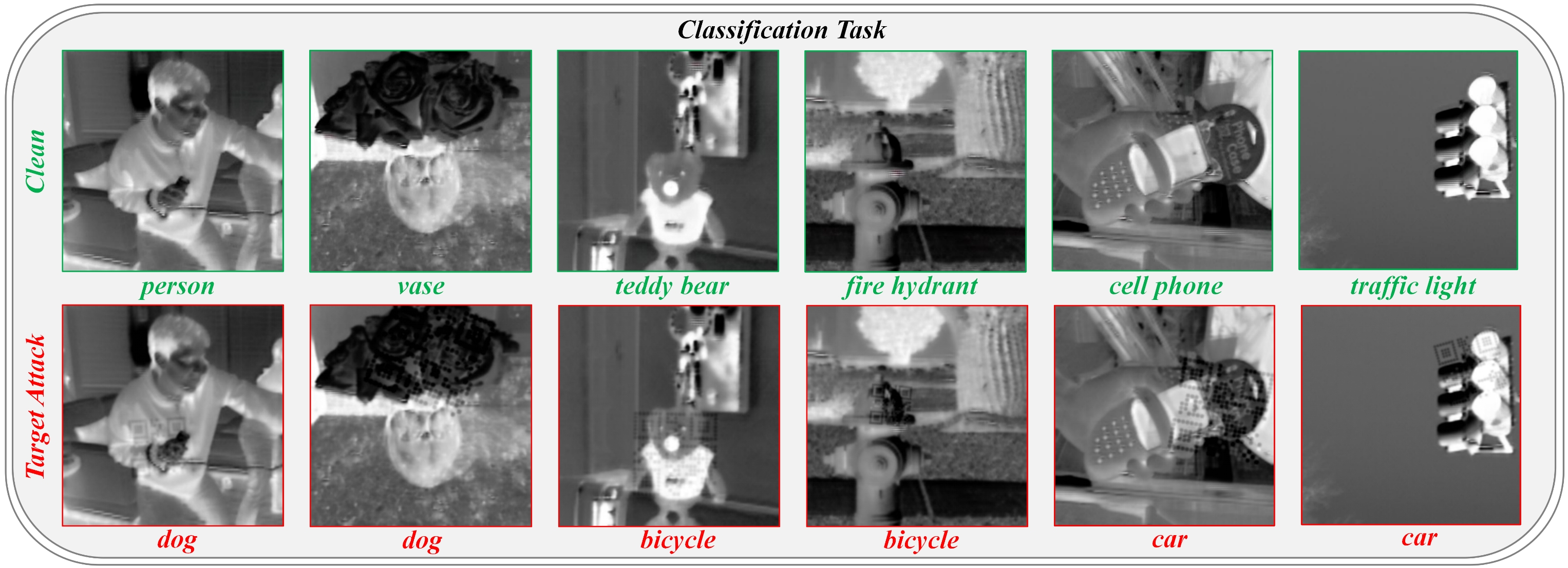}
\caption{Qualitative targeted-classification examples of \method. Clean infrared samples and their original predictions are shown above the corresponding adversarial samples. Green and red labels denote clean and attacker-specified target predictions, respectively.}
\label{fig:classification_examples}
\end{figure*}

\begin{table*}[t]
\centering
\small
\caption{Image-captioning robustness measured by clean-reference/source-preservation rate (\%). Lower values indicate larger semantic deviations.}
\label{tab:caption_transfer}
\resizebox{0.98\textwidth}{!}{
\begin{tabular}{llcccc}
\toprule
Image Encoder & Models
& AdvICRS
& HCB
& AdvGrid
& \method \\
\midrule
\multirow{4}{*}{\shortstack{OpenAI CLIP ViT-L/14}}
& LLaVA-1.5 (7B)
& 68.97 & 63.64 & 55.96 & \best{51.72} \\
& LLaVA-1.6 (7B)
& 63.52 & 50.00 & 35.78 & \best{29.56} \\
& OpenFlamingo (3B)
& 65.45 & 57.32 & 52.40 & \best{50.74} \\
& BLIP-2 FlanT5XL ViT-L (3.4B)
& 72.10 & 69.46 & 67.89 & \best{67.00} \\
\midrule
\multirow{2}{*}{\shortstack{EVA-CLIP ViT-G/14}}
& BLIP-2 FlanT5XL (4.1B)
& 61.24 & 58.79 & 47.56 & \best{41.65} \\
& InstructBLIP FlanT5XL (4.1B)
& 67.58 & 62.36 & 54.30 & \best{52.40} \\
\bottomrule
\end{tabular}}
\end{table*}

\begin{figure*}[t]
\centering
\includegraphics[width=0.98\textwidth]{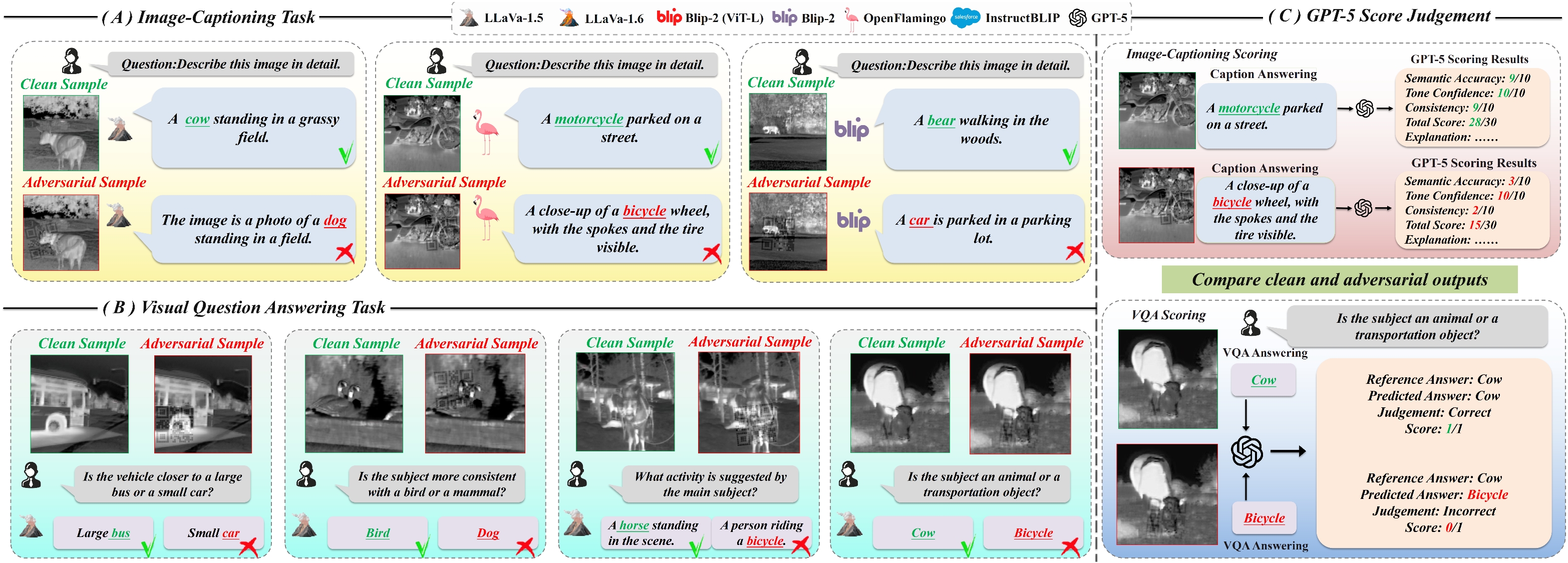}
\caption{Qualitative transfer examples of \method on image captioning and VQA. (A) Adversarial samples induce target-consistent changes in generated captions. (B) Target-agnostic VQA answers are redirected toward attacker-specified concepts. (C) GPT-5-based evaluation compares clean and adversarial outputs against clean-reference semantics.}
\label{fig:caption_vqa_examples}
\end{figure*}

\subsection{Progressive Black-Box Optimization}

The search space of \method is mixed discrete--continuous: trigger
geometry and rendering parameters are continuous, whereas the module
states are discrete. Jointly optimizing all variables creates a
high-dimensional and unstable black-box problem. We therefore adopt
a progressive strategy that proceeds from coarse spatial
configurations to fine-grained module topology.

At stage $k\in\{1,2,3\}$, \method optimizes a stage-specific variable
set $\Omega_k$:
\begin{equation}
\theta^{(k)\star}
=
\arg\min_{\theta\in\Omega_k}
\mathcal{L}_{\method}
\big(
A_{\mathrm{QR}}(x_i;\theta),s_i,t
\big).
\label{eq:stage_obj}
\end{equation}
The best solution from each stage initializes the next stage.

\textbf{Stage 1: coarse zone search.}
We partition the editable region into coarse spatial zones, with all
modules in each zone sharing a relaxed state. This stage optimizes
trigger position, scale, rotation, intensity, and zone-level
cold/hot tendencies to identify a promising placement and coarse
topology.

\textbf{Stage 2: module topology search.}
Starting from the best coarse solution, each editable module is
assigned an independent relaxed state. The optimizer then searches
for a fine-grained target-specific topology, which provides the main
semantic capacity of the trigger.

\textbf{Stage 3: rendering refinement.}
Given the topology obtained in Stage 2, we refine intensity, blur,
roundness, and small geometric corrections. This stage suppresses
sharp artifacts and improves the trade-off between target alignment
and visual similarity.

\textbf{Greedy module-flip refinement.}
Because relaxed optimization may not identify the best ternary
configuration, we perform a final local search over the decoded
module states. At iteration $\ell$, a candidate
$\widetilde{M}^{(\ell)}$ is generated by changing a small subset of
editable modules among cold, neutral, and hot states. The candidate
is accepted only if it decreases the complete objective:
\begin{equation}
M^{(\ell+1)}
=
\begin{cases}
\widetilde{M}^{(\ell)}, & \Delta\mathcal{L}<0,\\
M^{(\ell)}, & \text{otherwise},
\end{cases}
\label{eq:greedy}
\end{equation}
where
\begin{equation}
\Delta\mathcal{L}
=
\mathcal{L}_{\method}
\big(
\widetilde{M}^{(\ell)}
\big)
-
\mathcal{L}_{\method}
\big(
M^{(\ell)}
\big).
\end{equation}
The refinement terminates when the query budget is exhausted or no
further improvement is found.

Overall, the progressive strategy separates three coupled decisions:
where the trigger is placed, which module topology encodes the target
signal, and how the trigger is rendered. This decomposition reduces
early-stage search complexity and stabilizes fine-grained topology
optimization.

\begin{table*}[t]
\centering
\small
\caption{VQA robustness. We report clean-reference/source-preservation rates (\%); lower values indicate larger answer deviations. Methods are ordered from smaller to larger overall deviation.}
\label{tab:vqa_transfer}
\resizebox{0.98\textwidth}{!}{
\begin{tabular}{llcccc}
\toprule
Image Encoder & Models
& AdvICRS
& HCB
& AdvGrid
& \method \\
\midrule
\multirow{4}{*}{\shortstack{OpenAI CLIP ViT-L/14}}
& LLaVA-1.5 (7B)
& 68.75 & 58.08 & 58.41 & \best{45.65} \\
& LLaVA-1.6 (7B)
& 64.20 & 61.62 & 51.07 & \best{37.27} \\
& OpenFlamingo (3B)
& 62.80 & 56.40 & 50.25 & \best{47.80} \\
& BLIP-2 FlanT5XL ViT-L (3.4B)
& 83.33 & 75.20 & 67.89 & \best{66.17} \\
\midrule
\multirow{2}{*}{\shortstack{EVA-CLIP ViT-G/14}}
& BLIP-2 FlanT5XL (4.1B)
& 57.86 & 54.72 & 43.95 & \best{35.80} \\
& InstructBLIP FlanT5XL (4.1B)
& 63.20 & 58.75 & 49.40 & \best{41.60} \\
\bottomrule
\end{tabular}}
\end{table*}

\begin{figure*}[t]
\centering
\includegraphics[width=\textwidth]{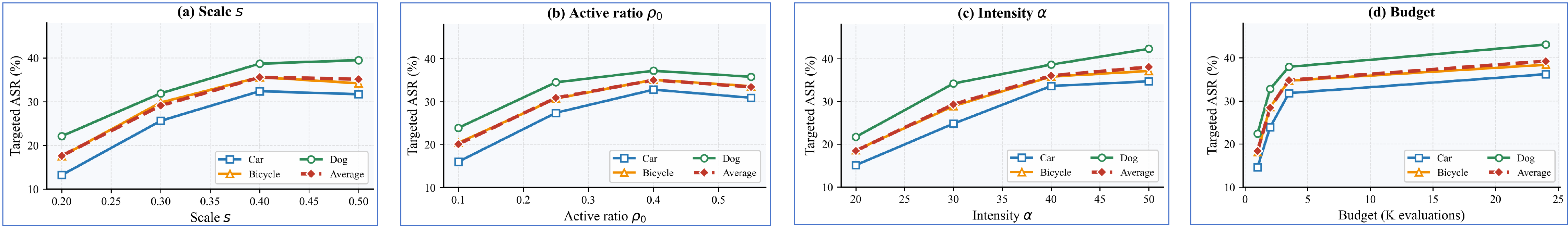}
\caption{Hyperparameter sensitivity of \method\ across trigger scale, active ratio, thermal intensity, and optimization budget, measured by ASR.}
\label{fig:hyper}
\end{figure*}

\begin{figure}[t]
\centering
\includegraphics[width=\linewidth]{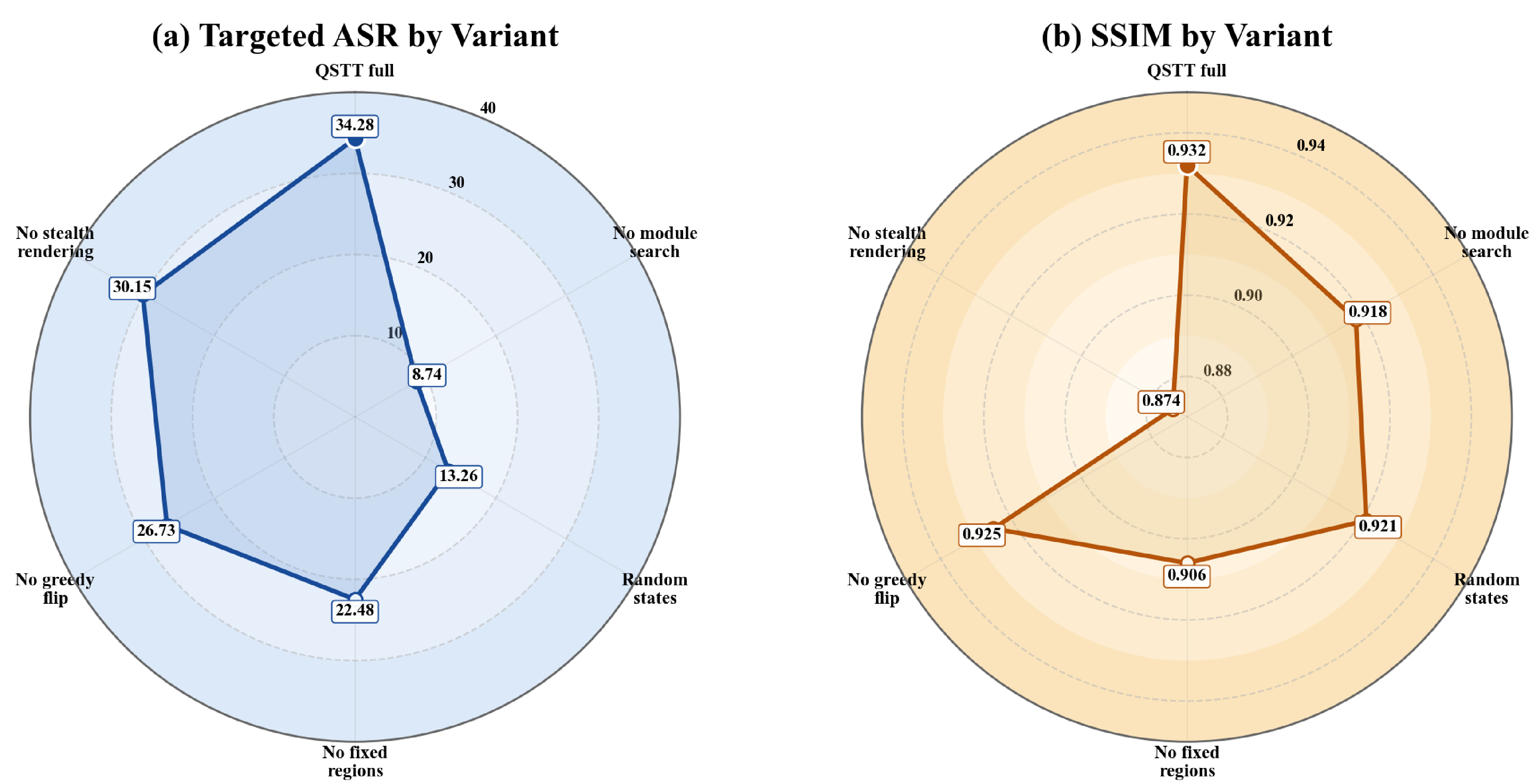}
\caption{Component ablation of \method. (a) Targeted ASR for the full method and its ablated variants. Removing module-topology search causes the largest degradation in attack effectiveness. (b) SSIM for the same variants. Disabling stealth rendering produces the largest reduction in visual similarity. Overall, the full \method achieves the best balance between targeted ASR and SSIM.}
\label{fig:ablation}
\end{figure}

\section{Experiments}

\subsection{Experimental Setup}

\textbf{Data and models.}
Following \citet{jiang2024infraredllava}, all deployed VLMs are infrared-adapted for thermal inputs. We use a 30-class infrared test set (10 images per class) and only keep samples correctly predicted by frozen CLIP encoders. Each sample is optimized for three targets: dog, car, and bicycle.
We adopt four CLIP backbones: OpenCLIP ViT-B/16~\citep{cherti2023openclip}, Meta-CLIP ViT-L/14~\citep{xu2024metaclip}, EVA-CLIP ViT-G/14~\citep{sun2023evaclip}, OpenAI CLIP ViT-L/14~\citep{radford2021clip}. For captioning and VQA transfer, we utilize six infrared-tuned generative IR-VLMs~\citep{liu2023llava,liu2024llavanext,awadalla2023openflamingo,li2023blip2,dai2023instructblip}.

\textbf{Attack protocol and evaluation.}
We run black-box zero-shot targeted attacks. Attack success is defined as the top-1 CLIP prediction of $x_i^{adv}$ matching target $t$. The main evaluation metrics are targeted attack success rate (ASR) and SSIM~\citep{wang2004ssim}.
Adversarial infrared images are directly tested on captioning/VQA without extra fine-tuning. Consistent target-biased outputs demonstrate that CLIP-side embedding shifts transfer to generation tasks. We measure semantic drift using GPT-5~\citep{openai2025gpt5} as the evaluator under the LLM-as-a-judge protocol~\citep{zheng2023judging}.

\textbf{Implementation Details.}
\method performs three-stage gradient-free black-box optimization over structured thermal QR modules, including zone exploration, module refinement, and stealth rendering. We use a version-1 QR carrier ($21\times21$ modules) with fixed functional regions and optimize only editable modules. Under the default fast setting, the three stages use population sizes of 40/50/60 and 12/16/20 generations. Thermal carriers are rendered at $224\times224$ resolution with rounded module kernels, while the objective jointly considers target alignment and visual stealth. All experiments are conducted on NVIDIA RTX 4090 GPUs.

\textbf{Baselines.}
We adapt three representative structured infrared attacks into the same source-to-target setting. \textbf{AdvGrid} uses grid-based thermal perturbations~\citep{tiliwalidi2025advgrid}, \textbf{AdvICRS} employs spline-based carriers~\citep{jia2025advicrs}, and \textbf{HCB} adopts hot/cold block patterns~\citep{hu2023hotcold}. All baselines follow the same clean filtering, target labels, query constraints, and downstream evaluation protocol for fair comparison.

\subsection{Targeted Classification Evaluation}

Table~\ref{tab:main} reports targeted ASR across four CLIP-style encoders and three target labels. \method achieves the best performance on all 12 backbone--target pairs, with a macro-average ASR of 34.28\%, substantially outperforming AdvGrid, HCB, and AdvICRS. Under the same clean-correct filtering, target labels, and black-box protocol, these results indicate that the optimized QR-module topology provides a more effective targeted steering space than grid-, curve-, or block-based thermal carriers. Attackability varies across backbones and targets: EVA-CLIP ViT-G/14 is comparatively more robust, whereas OpenCLIP ViT-B/16 and OpenAI CLIP ViT-L/14 are more vulnerable. Figure~\ref{fig:classification_examples} further shows that diverse clean infrared inputs can be redirected toward attacker-specified concepts such as dog, bicycle, and car, consistent with the quantitative results.

\subsection{Image Captioning and VQA Robustness}

We further evaluate whether adversarial images optimized only for targeted zero-shot classification transfer to downstream generation tasks without task-specific optimization. For image captioning, Table~\ref{tab:caption_transfer} reports clean-reference/source-preservation rates, where lower values indicate greater semantic deviation. \method achieves the lowest rates across all evaluated models, showing that QR-structured triggers affect both classification and generated descriptions. Figure~\ref{fig:caption_vqa_examples}(A) presents representative target-consistent caption shifts while preserving the overall infrared scene. Similar results are observed for VQA in Table~\ref{tab:vqa_transfer}, where \method again yields the lowest preservation rates. Since the questions are target-agnostic and the downstream models are excluded from attack optimization, these findings indicate that the induced embedding shift transfers beyond classification to generation-level reasoning. Figures~\ref{fig:caption_vqa_examples}(B) and (C) show representative target-biased answers and the GPT-5-based evaluation protocol, respectively.

\subsection{Ablation Study}

\subsubsection{Hyperparameter Sensitivity.}
We analyze four key hyperparameters, including trigger scale, active-module ratio, thermal intensity, and query budget, to assess their effects on targeted attack performance. As shown in Figure~\ref{fig:hyper}, increasing the trigger scale provides greater editable capacity and improves ASR, although the gain gradually saturates at larger scales. A similar trend is observed for the active-module ratio, where moderate activation enables stronger semantic steering than overly sparse configurations. Higher thermal intensity also improves ASR, with diminishing gains at stronger settings. For the query-budget analysis, we vary the total number of black-box evaluations across the complete optimization pipeline. QR-STT achieves substantial gains under small budgets, while the improvement gradually saturates as additional evaluations mainly refine the solution.

\subsubsection{Component Ablation of \method.}
We remove one component at a time while keeping the remaining attack pipeline unchanged, as shown in Figure~\ref{fig:ablation}. Removing fixed QR regions evaluates the contribution of the structural scaffold, disabling module-topology search tests the necessity of target-specific cold, neutral, and hot layouts, removing greedy module flipping measures the benefit of local discrete refinement, and disabling stealth rendering removes blur, low contrast, and rounded module boundaries. Figure~\ref{fig:ablation}(a) shows that module-topology search contributes most to targeted attack effectiveness, while Figure~\ref{fig:ablation}(b) shows that stealth rendering is most important for preserving visual similarity. The full \method achieves the best overall ASR--SSIM trade-off.

\section{Conclusion and Discussion}
We propose \method, a training-free black-box framework for targeted semantic steering of infrared VLMs. By preserving the fixed QR scaffold and optimizing editable thermal modules through progressive mixed discrete--continuous search, \method achieves strong attack effectiveness and cross-task transferability across classification, captioning, and VQA. Ablation studies validate the roles of topology search, discrete refinement, and stealth-oriented rendering, establishing QR-structured perturbations as an effective and interpretable attack space for infrared VLMs.

\bibliography{aaai2027}

\end{document}